\documentclass[sigconf]{acmart}
\AtBeginDocument{%
  }

\setcopyright{acmlicensed}
\copyrightyear{2026}
\acmYear{2026}
\setcopyright{cc}
\setcctype{by}
\acmConference[MM '26]{Proceedings of the 34th ACM International Conference on Multimedia}{November 10--14, 2026}{Rio de Janeiro, Brazil}
\acmBooktitle{Proceedings of the 34th ACM International Conference on Multimedia (MM '26), November 10--14, 2026, Rio de Janeiro, Brazil}
\acmDOI{10.1145/3767308.3836571}
\acmISBN{979-8-4007-2213-4/2026/11}

\usepackage[ruled,vlined,linesnumbered]{algorithm2e}
\usepackage{booktabs}
\usepackage{graphicx}
\usepackage{makecell}
\usepackage{multirow}
\usepackage{subcaption}
\usepackage[table]{xcolor}

\begin{document}

\title{LearnPrunerV2: Learnable Progressive Token Pruning for Vision-Language Models}

\author{Rinyoichi Takezoe}
\email{takezoerinyoichi@lixiang.com}
\orcid{0009-0000-1429-0289}
\affiliation{%
  \institution{Li Auto Inc.}
  \city{Beijing}
  \country{China}}

\author{Yaqian Li}
\correspondingauthor
\email{liyaqian@lixiang.com}
\orcid{0009-0002-7313-8679}
\affiliation{%
  \institution{Li Auto Inc.}
  \city{Shanghai}
  \country{China}}

\author{Weiren Wang}
\email{wangweiren@lixiang.com}
\orcid{0009-0007-6536-973X}
\affiliation{%
  \institution{Li Auto Inc.}
  \city{Beijing}
  \country{China}}

\author{Zihao Bo}
\email{bozihao1@lixiang.com}
\orcid{0000-0002-1391-9142}
\affiliation{%
  \institution{Li Auto Inc.}
  \city{Beijing}
  \country{China}}

\author{Huihong Shi}
\email{shihuihong@lixiang.com}
\orcid{0000-0002-7845-0154}
\affiliation{%
  \institution{Li Auto Inc.}
  \city{Shanghai}
  \country{China}}

\author{Anzhou Hou}
\email{houanzhou@lixiang.com}
\orcid{0009-0008-5899-2942}
\affiliation{%
  \institution{Li Auto Inc.}
  \city{Shanghai}
  \country{China}}

\author{Mo Guang}
\email{guangmo@lixiang.com}
\orcid{0009-0007-5246-9280}
\affiliation{%
  \institution{Li Auto Inc.}
  \city{Hangzhou}
  \country{China}}

\author{Jiale Yan}
\email{yanjiale@lixiang.com}
\orcid{0000-0003-4972-6315}
\affiliation{%
  \institution{Li Auto Inc.}
  \city{Shanghai}
  \country{China}}

\author{Kaiwen Long}
\email{longkaiwen@lixiang.com}
\orcid{0009-0000-7183-916X}
\affiliation{%
  \institution{Li Auto Inc.}
  \city{Shanghai}
  \country{China}}

\renewcommand{\shortauthors}{Rinyoichi Takezoe et al.}

\begin{abstract}
    Vision-Language Models (VLMs) have demonstrated remarkable multimodal capabilities but suffer from prohibitive computational overhead due to the massive number of visual tokens. While token pruning is a promising solution, existing methods still suffer from two major drawbacks: 1) unreliable pruning criteria that fail to accurately localize task-specific regions due to inherent attention biases or query-agnostic designs, and 2) suboptimal pruning strategies that employ rigid, non-adaptive schedules, overlooking the dynamic evolution of token redundancy across different model depths. To address these challenges, we propose LearnPrunerV2, which extends the previous two-stage heuristic framework (LearnPruner) into a fully learnable architecture. Specifically, we integrate lightweight learnable modules into every layer of the VLM backbone, which estimate token importance by fusing intrinsic visual saliency with extrinsic query relevance captured via a shadow token. By uncovering a consistent staircase redundancy pattern across VLM layers, we further introduce Budget-Deterministic Pruning (BDP) to establish a hardware-friendly schedule with fixed per-layer budgets. Extensive experimental results across various benchmarks and VLMs demonstrate that LearnPrunerV2 consistently outperform previous state-of-the-art methods. Notably, LearnPrunerV2 preserves 99.8\% of the original performance when pruning 77.8\% of visual tokens on LLaVA-v1.5, achieving a superior accuracy-efficiency trade-off.
    

\end{abstract}

\ccsdesc[500]{Computing methodologies~Artificial intelligence}

\keywords{Efficient Inference, Vision-Language Models}


\maketitle

\section{Introduction}
\label{sec:intro}

Following the rapid evolution of Large Language Models (LLMs)~\cite{achiam2023gpt, liu2024deepseek, yang2025qwen3}, Vision-Language Models (VLMs) have extended these sophisticated reasoning capabilities to the multimodal domain, leading to significant advancements in visual understanding and grounding. Generally, VLMs employ a pretrained visual encoder to transform visual inputs into a sequence of tokens, which are subsequently concatenated with textual instruction tokens as input for the LLM. Visual tokens often dominate the input sequence, typically numbering in the hundreds and potentially expanding to thousands for high-resolution images or long videos. Since the computational complexity of self-attention scales quadratically with sequence length, such a massive influx of visual tokens drastically inflates inference latency, rendering VLMs impractical for real-time or resource-constrained deployment.

To alleviate these computational bottleneck, numerous studies have explored token pruning as a viable strategy for reducing the visual sequence length. Given the inherent redundancy in visual modality, prior research has demonstrated that a substantial proportion of visual tokens can be discarded with minimal impact on model performance. Fundamentally, the effectiveness of token pruning hinges on two core dimensions: how to prune (the selection criterion) and where to prune (the pruning schedule). In this paper, we argue that existing approaches remain suboptimal in both aspects.

\begin{figure*}[tbp]
    \centering
    \includegraphics[width=0.85\linewidth]{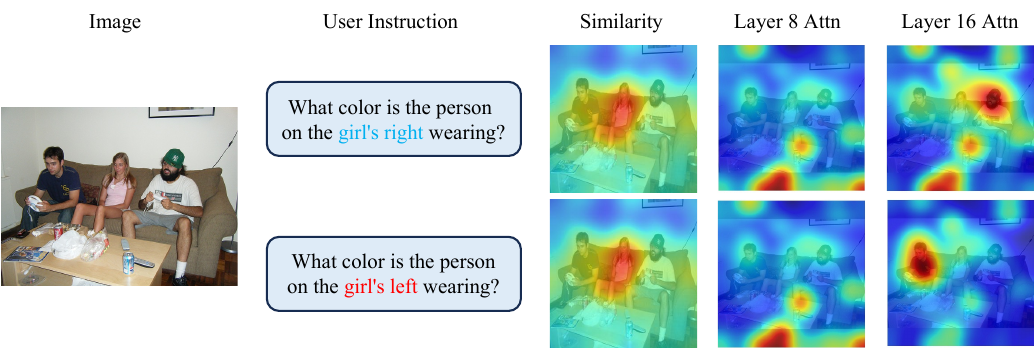}
    \caption{Visualization of cross-modal relevance scores in LLaVA-v1.5. 'Similarity' denotes the similarity scores between CLIP multimodal features, while 'Layer 8/16 Attn' represents the attention distribution of the last token within the LLM. Notably, CLIP features primarily capture isolated word-level semantics (e.g., the 'girl'); in contrast, intermediate LLM layers resolve global contextual semantics to accurately localize the intended target based on spatial relations.}
    \Description{Enjoying the baseball game from the third-base seats. Ichiro Suzuki preparing to bat.}
    \label{fig:motivation}
\end{figure*}

\textbf{Unreliability of Pruning Criteria.} Existing pruning criteria primarily fall into two categories: importance-based and diversity-based. The former utilizes attention scores from either the visual encoder or the LLM to identify significant tokens. However, as noted in LearnPruner~\cite{takezoelearnpruner}, visual encoders often suffer from attention sink that fail to localize foreground objects, while LLM attention exhibits a strong positional bias, with reliable query-aware signals only beginning to emerge from the intermediate layers onward. This creates a dilemma: early pruning is computationally efficient but "blind", while late-stage pruning is semantically guided but yields marginal speedup.
The latter category utilizes token-wise feature similarity to select the maximum coverage subset, aiming to preserve global visual diversity. These methods are inherently query-agnostic, making it challenging to adapt pruning results to different user instructions. While recent efforts~\cite{cai2025flashvlm, zhang2025cdpruner} attempt to incorporate auxiliary text encoders to facilitate cross-modal guidance, as shown in Figure~\ref{fig:motivation}, these encoders are limited to word-level granularity, failing to resolve the global semantic context of the image. This suggests that the internal attention of LLMs remains essential for capturing task-specific visual information.

\textbf{Suboptimality of Pruning Strategies.} Beyond selection criteria, existing methods typically employ static, one-pass pruning schedules or layer specific heuristics, overlooking the intrinsic coupling between token redundancy and VLM layer depth. Indeed, recent studies have revealed the layer-wise dynamics: AutoPrune~\cite{wang2025autoprune} demonstrates that VLM attention exhibits significant fluctuations across layers, particularly in reasoning-intensive scenarios. Furthermore, TwigVLM~\cite{shao2025growing} reveals that in the deep layers of VLMs, even the complete removal of visual tokens does not necessarily degrade performance. Collectively, these findings suggest that as visual information is incrementally integrated into linguistic representations, the significance of visual tokens is not uniform across layers but evolves throughout the hierarchy, motivating a progressive, layer-adaptive pruning strategy.

To address these limitations, LearnPruner~\cite{takezoelearnpruner} takes an initial step by adopting a two-stage token pruning framework that first removes redundant vision tokens via a learnable pruning module after the vision encoder, then retains only query-relevant tokens in the LLM's middle layers based on text-to-vision attention scores. While effective, its singular input signals and hand-crafted schedule cannot fully capture the continuous evolution of token redundancy.

Building upon this, we propose \textbf{LearnPrunerV2}, a purely learnable pruning framework for VLMs. Unlike the hybrid design of its predecessor, our framework adopts a unified architecture by integrating a lightweight pruning module into every layer of the VLM backbone. Moreover, a learnable "shadow token" is appended to the input sequence during the prefill stage to aggregate the preceding context and capture cross-modal correlations. Each pruning module subsequently combines intrinsic visual saliency from local features with extrinsic query relevance derived from the shadow token to predict importance scores for individual visual token. Through end-to-end joint optimization, LearnPrunerV2 aligns pruning decisions with the layer-specific evolution of visual significance, which in turn allows pruning modules to determine the optimal exit depth for each token. This dynamic mechanism enables the model to flexibly adjust its pruning trajectory based on the complexity of both images and queries. Furthermore, to address practical budget-constrained scenarios, we introduce Budget-Deterministic-Pruning (BDP) strategy. BDP utilizes offline-calibrated statistics to maintain a strictly fixed token count across every layer and adopts diversity-aware selection to mitigate the impact of a non-adaptive schedule. This ensures deterministic inference latency and maximized batching throughput, providing a robust solution for high-concurrency real-world deployment. 

Benefiting from its model-agnostic design, LearnPrunerV2 can be seamlessly integrated into diverse VLM architectures, such as LLaVA~\cite{liu2023visual,liu2024llavanext,lin2023video} and Qwen-VL~\cite{wang2024qwen2, Qwen2.5-VL}. Extensive experiments on multiple VQA benchmarks shows that our method outperforms prior state-of-the-art methods under different retention ratios. For example, when pruning 94.4\% visual tokens in LLaVA-NeXT-7B, LearnPrunerV2 preserve 96.2\% of the original model performance, while achieving a 9.7$\times$ reduction in TFLOPs. In summary, the contributions of our work are threefold:
\begin{itemize}
    \item We present LearnPrunerV2, a purely learnable token pruning framework that adopts a progressive pruning strategy. Compared to LearnPruner, it offers a more unified and flexible solution. 
    \item We introduce the Budget-Deterministic-Pruning schedule to assign fixed per-layer token budgets, enabling hardware-friendly inference.
    \item We conduct extensive experiments on various VLMs and vision-language benchmarks, demonstrating that LearnPrunerV2 consistently achieves state-of-the-art performance across different retention ratios.
\end{itemize} 






\section{Related Work}
\label{sec:related_work}

\subsection{Vision Language Models}
Building upon the foundational reasoning power of Large Language Models (LLMs)~\cite{touvron2023llamaopenefficientfoundation, jiang2023mistral7b, achiam2023gpt, liu2024deepseek, cai2024internlm2, yang2025qwen3}, Vision-Language Models (VLMs)~\cite{liu2023visual, li2024llavaonevisioneasyvisualtask, Qwen2.5-VL, zhu2025internvl3exploringadvancedtraining} demonstrate remarkable capabilities showing strong performance on diverse multimodal tasks such as image captioning, visual question answering, and visual grounding. A standard VLM typically comprises three components: a visual encoder to transform visual inputs into tokens, a vision-language connector to align the modality, and a LLM to serve as the cognitive engine for multimodal integration and complex reasoning. However, this paradigm incurs significant computational costs as the massive number of visual tokens. For instance, LLaVA-v1.5~\cite{liu2023visual} converts a 336x336 image into 576 tokens, advanced VLMs like LLaVA-NeXT~\cite{liu2024llavanext} (tiling-based) and Qwen2.5-VL~\cite{Qwen2.5-VL} (native dynamic resolution) support high-resolution inputs that further expand visual token sequences. Furthermore, in video understanding scenarios, the visual token count scales linearly with the number of sampled frames~\cite{lin2023video,longvila}. Such extensive sequences pose a severe bottleneck for real-time applications, driving the urgent need for efficient inference of VLMs.

\subsection{Vision Token Reduction for VLMs}
Given the inherent redundancy in visual signals, token reduction has emerged as a promising direction to enhance VLM efficiency. Various works~\cite{chu2024mobilevlm,cha2024honeybee,li2025tokenpacker,huang2024dynamic,wu2026hidrop} attempt to compress vision token into more compact representations. While effective in reducing computation, these methods typically change the model architecture and require retraining from scratch, resulting in high training costs and potential degradation of the performance upper bound. In this work, we mainly focus on token pruning techniques that can be directly applied to existing VLMs.

From the perspective of pruning criteria, existing methods can be broadly categorized into two types: importance-based and diversity-based. Importance-based methods~\cite{chen2024image,zhang2024sparsevlm,shao2025growing,zhang2025vispruner,yang2025visionzip,zhang2026d2pruner} utilize attention scores to select top-k important tokens and discard the rest. To remove query-irrelevant visual tokens, FastV~\cite{chen2024image} leverages the attention of the last text token at the second LLM layer as the principled signal, while SparseVLM~\cite{zhang2024sparsevlm} first identifies text tokens strongly correlated with visual input, and then employs their attention scores to achieve more precise guidance for pruning. Alternatively, to eliminate inherent visual redundancy, VisionZip~\cite{yang2025visionzip} and VisPruner~\cite{zhang2025vispruner} leverage [CLS] attention from the visual encoder to identify key salient tokens and perform pruning before the LLM. One limitation of importance-based methods is their reliance on attention maps, which makes them incompatible with Flash Attention~\cite{dao2022flashattention}. Diversity-based methods aims to select a diverse set based on similarity between tokens~\cite{alvar2025divprune,wen2025stop,zhang2025cdpruner,jeddi2025similarityawaretokenpruningvlm}. DivPrune~\cite{alvar2025divprune} formulates token pruning as a Max-Min Diversity Problem, whereas DART~\cite{wen2025stop} explicitly minimizes duplication by comparing candidate tokens against a set of pivots. Furthermore, CDPruner~\cite{zhang2025cdpruner} calculates inter-token similarity conditioned on their relevance to the current instruction and use the determinantal point process to maximize the conditional diversity.

From the perspective of pruning schedules, most studies conduct static pruning at a specific layer, overlooking the layer-wise evolution of token redundancy. Alternatively, some works adopt multi-stage pruning schedules~\cite{xing2024pyramiddrop,ye2025atp,jiang2025visa,wang2025autoprune,takezoelearnpruner}. For instance, PDrop~\cite{xing2024pyramiddrop} applies progressive pruning with uniform ratios and intervals; AutoPrune~\cite{wang2025autoprune} derives sample-specific policies by quantifying cross-modal mutual information; LearnPruner~\cite{takezoelearnpruner} employs a two-stage scheme, applying stage-specific pruning criteria to different phases. Despite these advancements, such heuristic designs remain misaligned with the dynamic nature of visual information processing, leading to suboptimal pruning schedules.

\section{LearnPrunerV2}
\label{sec:method}

\subsection{Preliminaries}
Existing VLMs generally consist of three primary components: a vision encoder, projector and LLM. Since our methods operate within the LLM, we first provide a formal definition of its architecture. The LLM is composed of a stack of Transformer-based~\cite{vaswani2017attention} decoder layers, each containing a causal self-attention to capture contextual dependencies and a feed-forward network for feature refinement. 

Given image and user instruction, inputs are discretized via vision encoder and tokenizer into $X_{vis}$ and $X_{text}$ respectively. These are then concatenated with system prompts to form the LLM input sequence $X = [X_{sys}, X_{vis}, X_{text}] \in \mathbb{R}^{N \times d}$, where $N$ is the sequence length and $d$ is the hidden dimension. In each decoder layer, the input $X$ is first projected into query $Q$, key $K$, and value $V$:
\begin{equation}
    Q = XW_Q, \quad K = XW_K, \quad V = XW_V
\end{equation}
where $W_Q, W_K, W_V \in \mathbb{R}^{d \times d}$ are projection matrices. The causal self-attention is then computed as:
\begin{equation}
    A = \text{softmax}\left(\frac{QK^T + M}{\sqrt{d}}\right), \quad O = AV
\end{equation}
where $M \in \mathbb{R}^{N \times N}$ is an upper triangular matrix to ensure each token can only attend to itself and previous tokens. Subsequently, the attention output is fed into the feed-forward network for non-linear transformation. Residual connections are applied to both the attention and ffn blocks to produce the output of current layer. After passing through all decoder layers, the final hidden states are utilized for next-token prediction.

\begin{figure*}[tbp]
    \centering
    \includegraphics[width=0.85\linewidth]{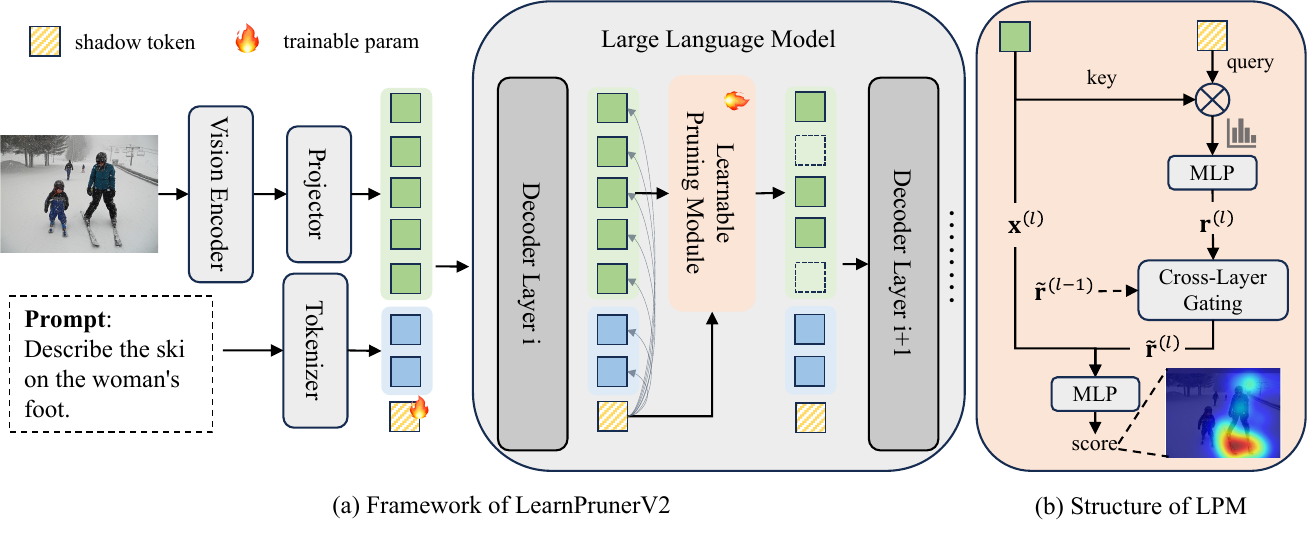}
    \caption{Overview of LearnPrunerV2. (a) Framework illustration, a learnable pruning module (LPM) is inserted into each decoder layer to evaluate token importance, and a learnable shadow token is appended after the user input to aggregate semantic context and capture cross-modal relevance. (b) Internal structure of the LPM. The visual hidden state is concatenated with the query-aware feature derived from the shadow token to predict the final importance score.}
    \Description{Enjoying the baseball game from the third-base seats. Ichiro Suzuki preparing to bat.}
    \label{fig:framework}
\end{figure*}

\subsection{Learnable Pruning Criterion}
Unlike prior works that rely on pre-defined heuristic rules, our approach employs learnable modules to evaluate token importance in a data-driven manner. The core challenge lies in constructing effective inputs for the module. A straightforward baseline is to use raw token features, similar to the Mixture-of-Depth (MoD) approach~\cite{raposo2024mixture}. However, while these features contain rich semantic information, they are inherently query-agnostic, capturing only intrinsic visual saliency rather than task-specific importance. Although text-to-vision attention provides such relevance, it faces two critical limitations. First, the requirement for explicit attention scores is incompatible with efficient attention implementations such as FlashAttention. Second, as observed in FastV~\cite{chen2024image}, attention scores in VLMs are heavily dominated by the textual modality, visual tokens often receive scores orders of magnitude lower (e.g., around $10^{-4}$ in LLaVA-1.5). Such faint signals lack sufficient numerical contrast, making it difficult for learnable modules to capture fine-grained importance differences.

To address these issues, we introduce a learnable shadow token, appended after the user instruction to aggregate semantic context and locate query-relevant visual regions. Specifically, the shadow token is initialized with the newline character, exploiting its natural capacity for summarizing preceding information in LLMs. During training, the input sequence is expanded to $[X_{sys}, X_{vis}, X_{ins}, $ $ X_{shd}, X_{ans}]$. Crucially, the shadow token can attend to all preceding tokens, but remains invisible to all other tokens via a customized causal mask, avoiding any perturbation to the model's original predictions. Within the decoder layer, we construct a relevance vector $\mathbf{a}_i \in \mathbb{R}^H$ for $i$-th visual token by computing the multi-head interaction between the shadow query $q_{shd}$ and corresponding visual keys $k_i \in K_{vis}$:
\begin{equation}
    \mathbf{a}_i = \left[ \frac{q_{shd}^{h} (k_i^{h})^T}{\sqrt{d}} \right]_{h=1}^H,
\end{equation}
where $H$ is the number of attention heads. Notably, this standalone query-key interaction avoids extracting from the internal attention matrix, ensuring compatibility with FlashAttention, and the use of pre-softmax logits preserves sufficient numerical contrast for optimization. The vector $\mathbf{a}_i$ is then mapped through a layer-specific MLP, denoted as $\text{MLP}_{\phi}$, to yield an enriched query-relevant feature $\mathbf{r}_i = \text{MLP}_{\phi}(\mathbf{a}_i) \in \mathbb{R}^{d_r}$. 

While this feature captures the cross-modal correlation at a single layer, prior work~\cite{wang2025autoprune} has shown that text-to-vision attention often exhibits pronounced inter-layer saliency fluctuations, particularly in complex scenarios. Directly relying on such unstable signals risks producing inconsistent pruning decisions. To stabilize importance estimation, we introduce a gating mechanism that incorporates historical information from the preceding layer. Concretely, given the current relevance feature $\mathbf{r}_i^{(l)}$ and its smoothed counterpart from the preceding layer $\tilde{\mathbf{r}}_i^{(l-1)}$, the gating weight is computed as:
\begin{equation}
    w_i^{(l)} = \sigma \left( \text{Linear}([\mathbf{r}_i^{(l)}, \tilde{\mathbf{r}}_i^{(l-1)}]) \right),
\end{equation}
where $\sigma$ is the sigmoid function and $[\cdot, \cdot]$ denotes concatenation. The smoothed relevance feature is then updated via:
\begin{equation}
    \tilde{\mathbf{r}}_i^{(l)} = w_i^{(l)} \cdot \mathbf{r}_i^{(l)} + (1 - w_i^{(l)}) \cdot \tilde{\mathbf{r}}_i^{(l-1)},
\end{equation}
with $\tilde{\mathbf{r}}_i^{(0)}=\mathbf{0}$. Intuitively, the gate learns to suppress transient attention noise while emphasizing persistent, instruction-aligned visual regions. 

Finally, to combine extrinsic query relevance with intrinsic visual saliency, $\tilde{\mathbf{r}}_i^{(l)}$ is concatenated with the token's hidden state $X_i^{(l)}$ and then fed into a shared scoring network $\text{MLP}_{\psi}$ to predict the importance score:
\begin{equation}
    s_{i}^{(l)} = \sigma \left( \text{MLP}_{\psi} \left( \left[ X_i^{(l)}, \tilde{\mathbf{r}}_i^{(l)} \right] \right) \right),
\end{equation}
where $s_{i}^{(l)}$ denotes the importance score of the $i$-th visual token at the$l$-th layer. By keeping $\text{MLP}_{\phi}$ layer-specific while sharing $\text{MLP}_{\psi}$ across all layers, we accommodate hierarchical semantic shifts while stabilizing the optimization process.

Since pruning requires discrete keep-or-drop decisions that are inherently non-differentiable, we employ the Straight-Through Estimator (STE)~\cite{bengio2013estimating} to bridge this gap. A binary mask is obtained by thresholding:
\begin{equation}
    \mathbf{M}_{i}^{(l)} = \text{STE} \left( \mathbb{I}(s_{i}^{(l)} > \tau) \right),
\end{equation}
where $\mathbb{I}(\cdot)$ is the indicator function and $\tau$ is set to 0.5. During the forward pass, $\mathbf{M}_{i}^{(l)} \in \{0, 1\}$ determines whether the token is kept or pruned; during the backward pass, the gradient bypasses the hard threshold and is approximated by that of the continuous score $s_{i}^{(l)}$, enabling end-to-end optimization. 
To ensure hierarchical consistency, we maintain a cumulative mask $\hat{\mathbf{M}}^{(l)}$ via a recursive Hadamard product:
\begin{equation}
    \hat{\mathbf{M}}^{(l)} = \hat{\mathbf{M}}^{(l-1)} \odot \mathbf{M}^{(l)},
\end{equation}
where $\hat{\mathbf{M}}^{(0)} = \mathbf{1}$, so that once a token is pruned, it is permanently excluded from all subsequent layers. Following~\cite{rao2021dynamicvit}, we apply this mask during training to cut off interactions between pruned and retained tokens, simulating the actual pruning behavior at inference time.

Our primary goal is to encourage the model to use the minimum number of vision tokens required to produce the correct answer. To achieve this, we introduce a pruning loss $\mathcal{L}_{prune}$ to constrain the number of preserved tokens, which is defined as:
\begin{equation}
    \mathcal{L}_{prune} = \left( \frac{1}{L \cdot N_{vis}} \sum_{l=1}^{L} \sum_{i=1}^{N_{vis}} \hat{\mathbf{M}}_i^{(l)} - \rho \right)^2,
\end{equation}
where $L$ is the number of pruning layers, $N_{vis}$ is the total number of visual tokens and we set $\rho=0$ to enforce maximal sparsity. With this regularization term, the final training objective is formulated as:
\begin{equation}
\mathcal{L} = \mathcal{L}_{ntp} + \lambda \mathcal{L}_{prune},
\end{equation}
where $\mathcal{L}_{ntp}$ is the standard next-token prediction loss and we set $\lambda=1$ in all our experiments. Notably, the shadow token is excluded from the prediction loss, allowing it to focus entirely on capturing cross-modal relevance without being distracted by the language modeling objective.

\begin{figure}[tbp]
    \centering
    \includegraphics[width=\linewidth]{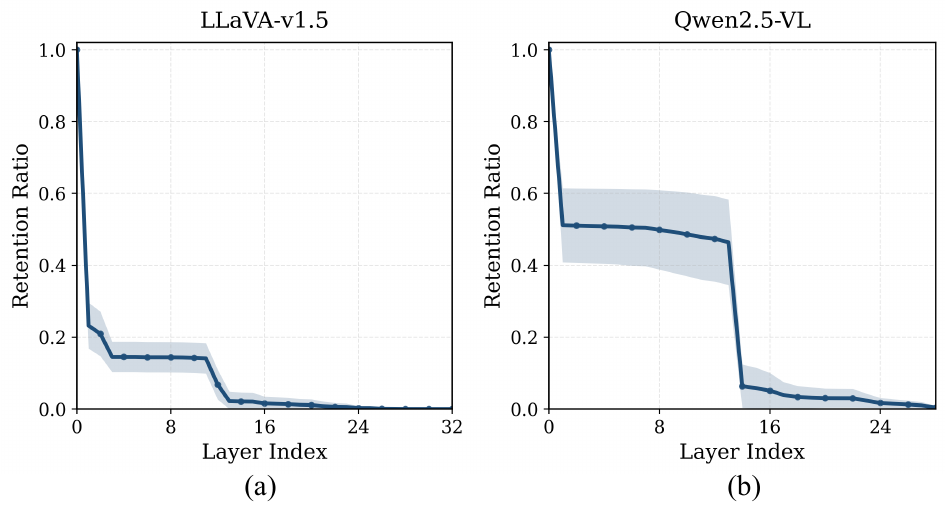}
    \caption{Statistics of learned pruning trajectories. We visualize the per-layer token retention ratio obtained via threshold-based pruning on LLaVA-v1.5 and Qwen2.5-VL. The solid lines denote the mean values, while the shaded regions represent the variance across 500 randomly sampled image-text pairs from LLaVA-mix665k.}
    \Description{Enjoying the baseball game from the third-base seats. Ichiro Suzuki preparing to bat.}
    \label{fig:retention_ratio}
\end{figure}

\subsection{Budget-Deterministic Pruning}
Once the pruning modules are trained, one can directly apply threshold-based pruning at each layer during inference, yielding sample-adaptive pruning schedules. However, practical deployment typically requires a fixed compute budget, which prior methods approximate by reporting the average number of retained tokens per layer. Furthermore, even with a fixed global budget, different inputs may produce distinct per-layer pruning trajectories, resulting in inconsistent memory access patterns and computation graphs, which is inherently hardware-unfriendly for optimized deployment.

To address these issues, we seek a deterministic per-layer schedule that fixes the token count at every layer. A natural question arises: how should such a schedule be determined? To this end, we investigate the underlying pruning characteristics learned by our modules. We randomly sample 500 image-text pairs from the LLaVA-mix665k dataset~\cite{liu2023visual} and execute threshold-based pruning with $\tau=0.5$ (identical to training). Figure~\ref{fig:retention_ratio} visualizes the mean and variance of the per-layer token retention ratio across all samples on LLaVA-v1.5 and Qwen2.5-VL. Surprisingly, the learned pruning trajectories exhibit high consistency across diverse inputs, with low inter-sample variance, and display a pronounced staircase pattern. This phenomenon suggests that through the joint optimization of all pruning modules, the model has autonomously identified a robust structural redundancy pattern in its architecture, reflecting the intrinsic stages of information flow and assimilation.

Taking LLaVA-v1.5 (Figure~\ref{fig:retention_ratio}(a)) as an example, the trajectory begins with a sharp drop at the first layer, immediately discarding uninformative visual content such as background regions. In the subsequent early layers, the retention ratio enters a stable plateau, where virtually no further pruning occurs, suggesting that the incremental information gained across these shallow layers provides little discriminative signal. As the model progresses into the middle layers, a second significant decline emerges. This aligns with prior findings~\cite{takezoelearnpruner,wu2026hidrop} where the middle layers serve as the primary stage for intensive cross-modal interaction. At this stage, visual features are progressively aligned with linguistic queries, causing a substantial portion of visual tokens to become redundant. Moving into the deep layers, the retention ratio gradually approaches zero, confirming that essential visual information has been integrated into the hidden states of the text tokens, rendering the original visual modality entirely redundant for final response generation.

\begin{table*}[t]
    \caption{Performance comparisons of different pruning methods on LLaVA-v1.5-7B. \textbf{Avg.} denotes the performance relative to the vanilla model, averaged across all benchmarks. Bold and underlined values denote the best and second-best results.}
    \centering
    \begin{tabular}{l|ccccccccc|c}
    \toprule
    \textbf{Method} & \textbf{GQA} & \textbf{SQA} & \textbf{VQA\textsuperscript{T}} & \textbf{POPE} & \textbf{MME} & \textbf{VQA\textsuperscript{v2}} & \textbf{MMB} & \textbf{MMB\textsuperscript{CN}} & \textbf{MMVet} & \textbf{Avg.} \\
    \rowcolor{gray!30}
    \multicolumn{11}{c}{\textit{Upper Bound, 576 Tokens (100\%)}} \\
    Vanilla & 61.9 & 69.5 & 58.2 & 85.9 & 1862 & 78.5 & 64.7 & 58.3 & 31.1 & 100.0\% \\
    \rowcolor{gray!30}
    \multicolumn{11}{c}{\textit{Retain Averaged 128 Tokens (\textcolor{blue}{$\downarrow$ 77.8\%})}} \\
    FastV & 49.6 & 60.2 & 50.6 & 59.6 & 1490 & 61.8 & 56.1 & 51.4 & 28.1 & 83.0\% \\
    SparseVLM & 56.0 & 67.1 & 54.9 & 80.5 & 1696 & 73.8 & 60.0 & 51.1 & 29.7 & 92.9\% \\
    VisPruner & 58.2 & \textbf{69.1} & 57.0 & 84.6 & 1794 & 75.8 & 62.7 & \underline{57.3} & \textbf{33.7} & 98.5\% \\
    DART & 57.9 & \textbf{69.1} & 56.3 & 80.4 & 1721 & 74.7 & 60.7 & \underline{57.3} & 30.9 & 95.8\% \\
    CDPruner & 59.9 & 69.0 & 56.2 & \textbf{87.7} & 1769.6 & 76.6 & 63.1 & 55.0 & \underline{32.8} & 98.3\% \\
    LearnPruner & \underline{60.3} & 68.5 & \underline{57.3} & \underline{86.7} & \underline{1820} & \underline{77.3} & \underline{63.8} & 56.8 & 30.8 & \underline{98.5\%} \\
    LearnPrunerV2 & \textbf{61.0} & 68.4 & \textbf{57.7} & 86.2 & \textbf{1843} & \textbf{77.7} & \textbf{64.4} & \textbf{58.5} & 32.4 & \textbf{99.8\%} \\
    \rowcolor{gray!30}
    \multicolumn{11}{c}{\textit{Retain Averaged 64 Tokens (\textcolor{blue}{$\downarrow$ 88.9\%})}} \\
    FastV & 46.1 & 51.1 & 47.8 & 48.0 & 1256 & 55.0 & 48.0 & 42.7 & 25.8 & 72.7\% \\
    SparseVLM & 52.7 & 62.2 & 51.8 & 75.1 & 1505 & 68.2 & 56.2 & 46.1 & 23.3 & 84.4\% \\
    VisPruner & 55.4 & \underline{69.1} & 55.8 & 80.4 & 1689 & 72.7 & 61.3 & 55.1 & \underline{32.3} & 95.0\% \\
    DART & 54.7 & \textbf{69.3} & 54.7 & 73.9 & 1705 & 71.3 & 59.5 & 54.0 & 26.5 & 91.1\% \\
    CDPruner & \underline{58.6} & 68.1 & 55.3 & \textbf{87.5} & 1721.5 & 75.4 & 61.1 & 53.2 & 30.5 & 95.8\% \\
    LearnPruner & \textbf{58.9} & 68.3 & \textbf{56.6} & \underline{86.8} & \underline{1750} & \underline{76.0} & \underline{62.6} & \textbf{55.7} & 29.6 & \underline{96.7\%} \\
    LearnPrunerV2 & \textbf{58.9} & 68.3 & \textbf{56.6} & 86.1 & \textbf{1830} & \textbf{76.3} & \textbf{63.3} & \underline{55.6} & \textbf{32.5} & \textbf{98.2\%} \\
    \rowcolor{gray!30}
    \multicolumn{11}{c}{\textit{Retain Averaged 32 Tokens (\textcolor{blue}{$\downarrow$ 94.4\%})}} \\
    FastV & 41.5 & 42.6 & 42.5 & 32.5 & 1090 & 43.4 & 37.8 & 33.2 & - & 52.0\% \\
    SparseVLM & 48.3 & 57.3 & 46.1 & 67.9 & 1290 & 58.6 & 51.4 & 40.6 & - & 68.0\% \\
    VisPruner & 52.2 & 69.2 & 53.9 & 72.7 & 1567 & 67.7 & 58.4 & 52.7 & \underline{28.8} & 89.4\% \\
    DART & 52.9 & \underline{69.3} & 52.2 & 69.1 & 1615 & 67.1 & 58.5 & 50.0 & 25.0 & 87.1\% \\
    CDPruner & 57.0 & \textbf{69.5} & 53.2 & \textbf{87.9} & \underline{1670} & 73.6 & 59.6 & 49.6 & 27.8 & 92.9\% \\
    LearnPruner & \underline{57.2} & 68.2 & \textbf{56.1} & \underline{84.5} & \underline{1672} & \underline{74.0} & \underline{60.8} & \textbf{55.5} & 26.1 & \underline{93.6\%} \\
    LearnPrunerV2 & \textbf{57.6} & 68.6 & \underline{55.7} & 85.8 & \textbf{1747} & \textbf{74.6} & \textbf{61.9} & \underline{54.4} & \textbf{29.9} & \textbf{95.7\%} \\
    \bottomrule
    \end{tabular}
    \label{tab:llava_v15}
\end{table*}

Motivated by the stability of the learned pruning trajectories, we propose Budget-Deterministic Pruning strategy, which repurposes the offline-calibrated average retention ratios $\boldsymbol{\eta} = [\eta_1, \eta_2, \dots, \eta_L]$ as a structural prior, constructing a deterministic per-layer budget that preserves the learned layer-wise pruning distribution. Specifically, given a user-specified budget $\bar{n}$, defined as the average number of retained visual tokens per layer, the total token budget across all layers is $\bar{n} \cdot L$. Since the calibrated trajectory would naturally consume $N_{vis} \cdot \sum_{l=1}^{L} \eta_l$ tokens in total, we compute a global scaling factor to reconcile the two:
\begin{equation}
    \alpha = \frac{\bar{n} \cdot L}{N_{vis} \cdot \sum_{l=1}^{L} \eta_l}.
\end{equation}
The per-layer token budget is then determined as:
\begin{equation}
    n_l = \lfloor \eta_l \cdot N_{vis} \cdot \alpha \rceil, \quad l = 1, 2, \dots, L,
\end{equation}
where $\lfloor \cdot \rceil$ denotes rounding to the nearest integer. This formulation applies a uniform scaling to the calibrated pruning profile, ensuring that the relative layer-wise distribution is preserved under any target budget.

While deterministic budgets ensure computational predictability, they inevitably sacrifice instance-level adaptivity---a key advantage of threshold-based pruning. To mitigate this, we introduce a diversity-aware selection mechanism that dynamically optimizes the token composition within each layer's fixed capacity. At each layer $l$, we first identify a candidate set of informative tokens whose importance scores exceed the threshold $\tau$, with the candidate count denoted as $n'_l = \sum_{i=1}^{N_{vis}} \mathbb{I}(s_i^{(l)} > \tau)$. If $n'_l \leq n_l$, the budget is sufficient to accommodate the entire candidate set, and we simply retain the top-$n_l$ tokens by importance score. Conversely, when $n'_l > n_l$, the model considers more tokens important than the allocated budget permits. In this case, naively selecting the top-$n_l$ tokens by scores may lead to the loss of critical visual information. Instead, we ensure more comprehensive visual representations by adapting the principle of diversity-aware selection from DART~\cite{wen2025stop}. Under this approach, the top-$n_p$ tokens ranked by importance score are first selected as pivot tokens. For each remaining candidate, we compute its cosine similarity to all pivots and identify its maximum similarity value. The remaining $n_l - n_p$ slots are filled by candidates with the lowest maximum similarity. Ultimately, this mechanism strikes a balance between importance and diversity, which is particularly effective in low-budget scenarios.

\section{Experiments}
\label{sec:experiments}

\begin{table*}[t!]
    \caption{Performance comparisons of different pruning methods on LLaVA-NeXT-7B.}
    \label{tab:llava_next}
    \vspace{-6pt}
    \centering
    \begin{tabular}{l|ccccccccc|c}
    \toprule
    \textbf{Method} & \textbf{GQA} & \textbf{SQA} & \textbf{VQA\textsuperscript{T}} & \textbf{POPE} & \textbf{MME} & \textbf{VQA\textsuperscript{v2}} & \textbf{MMB} & \textbf{MMB\textsuperscript{CN}} & \textbf{MMVet} & \textbf{Avg.} \\
    \rowcolor{gray!30}
    \multicolumn{11}{c}{\textit{Upper Bound, 2,880 Tokens (100\%)}} \\
    Vanilla & 64.2 & 70.2 & 61.3 & 86.3 & 1842 & 81.2 & 67.9 & 60.5 & 40.0 & 100.0\% \\
    \rowcolor{gray!30}
    \multicolumn{11}{c}{\textit{Retain Averaged 640 Tokens (\textcolor{blue}{$\downarrow$ 77.8\%})}} \\
    FastV & 62.0 & 69.1 & 60.0 & 79.5 & 1807 & 79.5 & 65.8 & 53.5 & 33.3 & 94.4\% \\
    SparseVLM & 60.3 & 67.7 & 57.8 & 85.3 & 1772 & 77.1 & 65.7 & 58.6 & 36.1 & 95.4\% \\
    DART & 61.3 & 68.2 & 59.5 & 85.0 & 1741 & 78.3 & 64.9 & 57.1 & 36.9 & 95.7\% \\
    CDPruner & 62.6 & 67.9 & 58.4 & \textbf{87.3} & 1808 & 79.9 & 66.3 & 57.5 & \textbf{38.8} & 97.4\% \\
    LearnPruner & \underline{63.8} & \underline{70.0} & \underline{60.5} & 85.1 & \underline{1837} & \underline{80.5} & \underline{67.5} & \underline{59.5} & 36.1 & \underline{98.1\%} \\
    LearnPrunerV2 & \textbf{64.2} & \textbf{70.1} & \textbf{60.6} & \underline{86.9} & \textbf{1856} & \textbf{81.2} & \textbf{68.0} & \textbf{60.7} & \underline{38.7} & \textbf{99.7\%} \\
    \rowcolor{gray!30}
    \multicolumn{11}{c}{\textit{Retain Averaged 320 Tokens (\textcolor{blue}{$\downarrow$ 88.9\%})}} \\
    FastV & 54.9 & 68.2 & 54.8 & 49.5 & 1539 & 69.6 & 60.0 & 42.5 & 20.0 & 78.6\% \\
    SparseVLM & 57.7 & 67.3 & 55.9 & 76.9 & 1694 & 73.4 & 64.3 & 56.7 & 32.8 & 91.0\% \\
    DART & 61.7 & 67.5 & 57.6 & 81.0 & 1720 & 75.7 & 64.2 & 58.2 & \underline{35.7} & 94.1\% \\
    CDPruner & 61.6 & 67.8 & 57.4 & \textbf{87.2} & 1758 & \underline{78.4} & 65.5 & 55.7 & 35.6 & 95.2\% \\
    LearnPruner & \underline{62.2} & \textbf{68.6} & \underline{58.4} & 84.8 & \textbf{1845} & 78.3 & \underline{66.8} & \underline{58.3} & 35.0 & \underline{96.3\%} \\
    LearnPrunerV2 & \textbf{63.5} & \underline{68.5} & \textbf{59.3} & \underline{87.0} & \underline{1844} & \textbf{80.0} & \textbf{67.8} & \textbf{60.2} & \textbf{36.6} & \textbf{98.2\%} \\
    \rowcolor{gray!30}
    \multicolumn{11}{c}{\textit{Retain Averaged 160 Tokens (\textcolor{blue}{$\downarrow$ 94.4\%})}} \\
    DART & 56.8 & \textbf{67.8} & 54.9 & 75.3 & 1626 & 72.5 & 62.0 & 53.6 & 32.2 & 88.9\% \\
    CDPruner & \underline{60.8} & 67.5 & \underline{55.4} & \textbf{86.8} & 1738 & \underline{76.7} & 64.2 & 53.8 & \underline{34.1} & \underline{93.3\%} \\
    LearnPruner & 58.7 & \underline{67.6} & 55.0 & 83.1 & \textbf{1784} & 76.2 & \underline{65.3} & \underline{55.8} & 32.8 & 92.8\% \\
    LearnPrunerV2 & \textbf{61.6} & \textbf{67.8} & \textbf{58.0} & \underline{85.6} & \underline{1781} & \textbf{77.9} & \textbf{65.5} & \textbf{59.2} & \textbf{37.1} & \textbf{96.2\%} \\
    \bottomrule
    \end{tabular}
\end{table*}

\subsection{Experimental Settings}

\textbf{Datasets and Benchmarks.} We conduct comprehensive evaluations on 9 widely-used image-based multimodal benchmarks, including GQA~\cite{hudson2019gqa}, ScienceQA-IMG~\cite{lu2022learn}, TextVQA~\cite{singh2019towards}, POPE~\cite{li2023evaluating}, MME~\cite{fu2023mme}, VQAv2~\cite{goyal2017making}, MMBench~\cite{liu2024mmbench}, MMBench-CN~\cite{liu2024mmbench} and MMVet~\cite{yu2023mm}. Moreover, we experiment with 3 video understanding benchmarks, including TGIF-QA~\cite{jang2017tgif}, MSVD-QA~\cite{xu2017video}, and MSRVTT-QA~\cite{xu2017video}. All experiments on these benchmarks follow the default settings and evaluation metrics.

\textbf{Implementation Details.} In the experiments, we keep the base VLM weights frozen and use the LLaVA-mix665k dataset~\cite{liu2023visual} to train our pruning modules. We apply our LearnPrunerV2 to various VLM architectures, including LLaVA-v1.5~\cite{liu2023visual} for image understanding, LLaVA-NeXT~\cite{liu2024llavanext} for high-resolution inputs, Video-LLaVA~\cite{lin2023video} for video scenarios (see Appendix for results) and Qwen2.5-VL~\cite{Qwen2.5-VL} 
for broader generalization. For comprehensive comparison, we select several recent token pruning methods as baselines, including importance-based methods (e.g., FastV~\cite{chen2024image}, SparseVLM~\cite{zhang2024sparsevlm} and VisPruner~\cite{zhang2025vispruner}), diversity-based approaches (e.g., DART~\cite{wen2025stop} and CDPruner~\cite{zhang2025cdpruner}), and LearnPruner~\cite{takezoelearnpruner} which introduces learnable parameters for pruning and employs a heuristic pruning schedule.

\begin{table}[t]
    \caption{Performance comparisons of different pruning methods on Qwen2.5-VL-7B.}
    \label{tab:qwen_vl}
    \vspace{-6pt}
    \centering
    \setlength{\tabcolsep}{3.5pt} 
    \begin{tabular}{l|ccccc|c}
    \toprule
    \textbf{Method} & \textbf{GQA} & \textbf{SQA} & \textbf{VQA\textsuperscript{T}} & \textbf{MME} & \textbf{MMB} & \textbf{Avg.} \\
    \rowcolor{gray!30}
    \multicolumn{7}{c}{\textit{Upper Bound, 1280 Tokens (100\%)}} \\
    Vanilla & 60.9 & 75.9 & 83.3 & 2314 & 83.8 & 100.0\% \\
    \rowcolor{gray!30}
    \multicolumn{7}{c}{\textit{Retain Averaged 426 Tokens (\textcolor{blue}{$\downarrow$ 66.7\%})}} \\
    FastV & 58.4 & \underline{79.1} & \textbf{81.5} & 2330 & 81.4 & 99.2\% \\
    LearnPruner & \underline{59.3} & 77.4 & \underline{81.1} & \underline{2339} & \underline{83.2} & \underline{99.4\%} \\
    LearnPrunerV2 & \textbf{60.0} & \textbf{81.4} & 80.7 & \textbf{2355} & \textbf{83.4} & \textbf{100.8\%} \\
    \rowcolor{gray!30}
    \multicolumn{7}{c}{\textit{Retain Averaged 284 Tokens (\textcolor{blue}{$\downarrow$ 77.8\%})}} \\
    FastV & 53.6 & \underline{77.8} & 79.2 & 2228 & 79.1 & 95.3\% \\
    LearnPruner & \underline{59.0} & 77.0 & \textbf{79.6} & \underline{2358} & \textbf{82.2} & \underline{98.8\%} \\
    LearnPrunerV2 & \textbf{59.1} & \textbf{82.2} & \textbf{79.6} & \textbf{2362} & \textbf{82.2} & \textbf{100.2\%} \\
    \rowcolor{gray!30}
    \multicolumn{7}{c}{\textit{Retain Averaged 142 Tokens (\textcolor{blue}{$\downarrow$ 88.9\%})}} \\
    FastV & 42.8 & 70.5 & 62.9 & 1565 & 56.2 & 74.7\% \\
    LearnPruner & \underline{51.9} & \underline{76.1} & \textbf{76.1} & \underline{2292} & \textbf{81.8} & \underline{94.7\%} \\
    LearnPrunerV2 & \textbf{56.6} & \textbf{82.2} & \underline{74.9} & \textbf{2320} & \underline{81.0} & \textbf{97.6\%} \\
    \bottomrule
    \end{tabular}
    \vspace{-12pt}
\end{table}

\subsection{Main Results}

\noindent\textbf{Results on LLaVA-v1.5.} As shown in Table~\ref{tab:llava_v15}, we first evaluate our method on LLaVA-v1.5-7B, where images are represented by a fixed sequence of 576 tokens. The results demonstrate that our method consistently outperforms all previous methods across all three retention ratios. Notably,  when retaining 128 tokens, our method maintains a remarkable 99.8\% of the vanilla model's performance, surpassing the previous state-of-the-art method by 1.3\%, and this performance gap further widens to $2.1\%$ when the budget is restricted to only 32 tokens. Since both our model and LearnPruner employ learnable modules, this significant lead directly validates the superiority of our proposed pruning criterion and pruning schedule.

\begin{table*}[!t]
    \caption{Ablation study on different configurations of the learnable pruning module (LPM).}
    \label{tab:ablations_lpm}
    \vspace{-6pt}
    \centering

    \begin{subtable}[t]{0.32\linewidth}
        \centering
        \caption{Ablation study on input components of the learnable pruning module.}
        \begin{tabular}{lc}
            \toprule
            \textbf{Input Components} & \textbf{Avg.} \\
            \midrule
            hidden states & 96.6 \\
            + relevance feature & 97.2 \\
            + gating mechanism (ours) & \textbf{97.8} \\
            \bottomrule
        \end{tabular}
    \end{subtable}
    \hfill 
    \begin{subtable}[t]{0.32\linewidth}
        \centering
        \caption{Ablation study on source of relevance feature.}
        \begin{tabular}{lc}
            \toprule
            \textbf{Relevance Source} & \textbf{Avg.} \\
            \midrule
            last token attention & 96.5 \\
            last token logit & 97.4 \\
            shadow token (ours) & \textbf{97.8} \\
            \bottomrule
        \end{tabular}
    \end{subtable}
    \hfill
    \begin{subtable}[t]{0.32\linewidth}
        \centering
        \caption{Ablation study on parameter sharing scheme of the learnable pruning module.}
        \begin{tabular}{lc}
            \toprule
            \textbf{Parameter Sharing} & \textbf{Avg.} \\
            \midrule
            fully independent & 97.3 \\
            fully shared & 96.9 \\
            hybrid (ours) & \textbf{97.8} \\
            \bottomrule
        \end{tabular}
    \end{subtable}
\end{table*}

\begin{table*}[!t]
    \centering
    \begin{minipage}[t]{0.48\textwidth}
        \captionof{table}{Ablation study on token selection strategy under Budget-Deterministic Pruning (BDP).}
        \label{tab:ablation_bdp}
        \vspace{-6pt}
        \centering
        \begin{tabular}{l|ccccc|c}
            \toprule
            \textbf{strategy} & \textbf{GQA} & \textbf{SQA} & \textbf{VQA\textsuperscript{T}} & \textbf{POPE} & \textbf{MME} & \textbf{Avg.} \\
            \rowcolor{gray!30}
            \multicolumn{7}{c}{\textit{Retain Averaged 64 Tokens}} \\
            top-k & \textbf{58.9} & \textbf{68.7} & \textbf{56.6} & 85.9 & 1805 & 97.7\% \\
            ours & \textbf{58.9} & 68.3 & \textbf{56.6} & \textbf{86.1} & \textbf{1830} & \textbf{97.8\%} \\
            \rowcolor{gray!30}
            \multicolumn{7}{c}{\textit{Retain Averaged 32 Tokens}} \\
            top-k & 56.7 & \textbf{68.7} & 55.2 & 85.1 & 1716 & 95.3\% \\
            ours & \textbf{57.6} & 68.6 & \textbf{55.7} & \textbf{85.8} & \textbf{1747} & \textbf{96.2\%} \\
            \bottomrule
        \end{tabular}
    \end{minipage}
    \hfill
    \begin{minipage}[t]{0.48\textwidth}
        \captionof{table}{Efficiency analysis on the POPE benchmark. All experiments are conducted on a single A100 GPU.}
        \label{tab:performance_metrics}
        \vspace{-6pt}
        \centering
        \setlength{\tabcolsep}{3.2pt}
        \resizebox{\linewidth}{!}{%
        \begin{tabular}{l|ccccc}
        \toprule
        \textbf{Method} & \textbf{Token} & \textbf{TFLOPs} & \textbf{Prefill (s)} & \textbf{Total (s)} & \textbf{\makecell{KV Cache\\(MB)}} \\
        \midrule
        LLaVA-NeXT & 2880 & 33.8 {\scriptsize(1.0$\times$)} & 3197.1 {\scriptsize(1.0$\times$)} & 3569.8 {\scriptsize(1.0$\times$)} & 1166.6 {\scriptsize(1.0$\times$)} \\
        \midrule
        \multirow{2}{*}{+ \textbf{ours}} & 640 & 10.2 {\scriptsize(3.3$\times$)} & 1231.7 {\scriptsize(2.6$\times$)} & 1783.3 {\scriptsize(2.0$\times$)} & 370.9 {\scriptsize(3.1$\times$)} \\
        & 160 & 3.5 {\scriptsize(9.7$\times$)} & 514.3 {\scriptsize(6.2$\times$)} & 967.5 {\scriptsize(3.7$\times$)} & 133.4 {\scriptsize(8.7$\times$)} \\
        \bottomrule
        \end{tabular}}
    \end{minipage}
\end{table*}

\noindent\textbf{Results on LLaVA-NeXT.} To verify its scalability for high-resolution scenarios, we evaluate our method on LLaVA-NeXT-7B. LLaVA-NeXT handles high-resolution inputs by partitioning the image into multiple sub-grids and encoding them independently, yielding a massive baseline of 2,880 visual tokens. As shown in Table~\ref{tab:llava_next}, LearnPrunerV2 retains 99.7\% of the original performance while utilizing only $22.2\%$ of the visual tokens. Furthermore, on several benchmarks such as POPE, MME, and MMBench, our pruned model achieves performance that is either comparable to or even superior to its vanilla counterpart. These results indicate that high-resolution inputs inherently contain substantial visual redundancy, where many tokens may contribute marginal or even distracting signals.
By removing uninformative visual context, the model can more accurately focus on the most salient features, thereby leading to more precise predictions.

\noindent\textbf{Results on Qwen2.5-VL.} To further assess the broad applicability of LearnPrunerV2, we extend our evaluation to Qwen2.5-VL-7B, a leading open-source VLM with an architecture distinct from the LLaVA family. Since the original SFT data for Qwen2.5-VL is not publicly accessible, we leverage the LLaVA-mix665k dataset to generate model responses, which serve as pseudo-labels for training our pruning modules. Furthermore, although Qwen2.5-VL naturally supports native dynamic resolution, we fix the input size to a standardized budget of 1,280 tokens to ensure a fair and consistent comparison across all pruning methods. As shown in Table~\ref{tab:qwen_vl}, LearnPrunerV2 still maintain its dominant performance, even exceeding 100\% average relative accuracy at $66.7\%$ and $77.8\%$ pruning rates, showing the robust generalizability of our approach.

\subsection{Ablation Study}
We evaluate the learnable pruning criterion and budget-deterministic pruning strategy on LLaVA-v1.5-7B. Unless otherwise specified, all experiments use a fixed budget of 64 tokens, and we report average relative performance over GQA, SQA, VQA\textsuperscript{T}, POPE, and MME.

\noindent\textbf{Input components.} As shown in Table~\ref{tab:ablations_lpm}(a), we compare three configurations for the pruning module's input. When relying solely on the hidden states, the module determines token importance based only on intrinsic visual redundancy, which results in a performance of $96.6\%$. By concatenating the relevance feature, the model can further incorporate cross-modal correlations with the textual query, boosting the average performance to $97.2\%$. However, inter-layer attention fluctuations often introduce distracting noise. The gating mechanism mitigates this by filtering transient signals, stabilizing importance estimation and yielding a performance of $97.8\%$.

\noindent\textbf{Relevance source.} Table~\ref{tab:ablations_lpm}(b) investigates the influence of different cross-modal relevance sources. While the last instruction token is a conventional choice for sequence aggregation, utilizing its attention scores results in poor performance ($96.5\%$) because the attention scores are numerically overwhelmed by the hidden states. Although last token logits ($97.4\%$) alleviate the scaling bottleneck, they still underperform the shadow token ($97.8\%$), which we attribute to the inherent multi-task interference between next-token prediction and pruning guidance. In contrast, the shadow token serves as a dedicated aggregator that focuses exclusively on capturing instruction-relevant token importance.

\noindent\textbf{Parameter sharing scheme.} The learnable pruning module mainly consists of a relevance projector ($\text{MLP}_{\phi}$) and a scoring network ($\text{MLP}_{\psi}$). Table~\ref{tab:ablations_lpm}(c) evaluates different parameter sharing schemes for these components. Our hybrid strategy achieves the best performance ($97.8\%$), outperforming both fully independent ($97.3\%$) and fully shared ($96.9\%$) counterparts. This is because attention logits exhibit significant inter-layer variance, necessitating a layer-specific $\text{MLP}_{\phi}$ for adaptive mapping. Conversely, a shared $\text{MLP}_{\psi}$ receives more gradients from different layers, facilitating robust convergence and stabilizing the optimization to learn a generalized pruning criterion.

\noindent\textbf{Visual token selection.} Table~\ref{tab:ablation_bdp} evaluates the impact of our diversity-aware selection against the standard Top-K strategy under a fixed layer-wise budget. While both approaches perform similarly at a higher budget of 64 tokens ($97.8\%$ vs. $97.7\%$), our strategy demonstrates a clear advantage when the budget is restricted to 32 tokens, outperforming Top-K by $0.9\%$. This gap suggests that simple Top-K selection is prone to information redundancy, as tokens with the highest scores are often spatially clustered or semantically similar. In contrast, our approach encourages the selection of more diverse visual cues, thereby broadening the overall information coverage within the restricted budget.

\subsection{Efficiency Analysis}
To demonstrate the efficiency of LearnPrunerV2, we measure the FLOPs, prefill time, total inference time and KV cache size of the pruned model compared to the vanilla LLaVA-NeXT-7B. All experiments are conducted on the POPE benchmark. As shown in Table~\ref{tab:performance_metrics}, when the number of visual tokens is reduced from 2,880 to 160, we achieve a 9.7$\times$ reduction in TFLOPs and a 6.2$\times$ speedup in prefill time. Furthermore, the KV cache size is slashed by 8.7$\times$, while the total inference time improves by 3.7$\times$. These results highlight the effectiveness of our framework in reducing the computational overhead of VLMs

\section{Conclusion}
\label{sec:conclusion}
We introduce LearnPrunerV2, a learnable progressive token pruning framework for accelerating VLMs. It estimates token importance from intrinsic visual saliency and query relevance, and employs a Budget-Deterministic Pruning (BDP) schedule that assigns fixed per-layer budgets by exploiting the inherent architectural redundancy of VLMs. Extensive experiments on multiple VLM architectures and benchmarks demonstrate strong accuracy-efficiency trade-offs and clear gains over prior methods.

\bibliographystyle{ACM-Reference-Format}
\bibliography{main}

\appendix
\clearpage
\section{Appendix}

\subsection{Results on Video-LLaVA}
To further assess the broad applicability of LearnPrunerV2, we extend our evaluation from the image domain to video understanding. Video-LLaVA~\cite{lin2023video} processes each video by uniformly sampling 8 frames and encoding them independently, producing a total of 2,048 visual tokens. We evaluate on three established video quesion answering benchmarks: TGIF-QA~\cite{jang2017tgif}, MSVD-QA~\cite{xu2017video} and MSRVTT-QA~\cite{xu2017video}. Following~\cite{zhang2025vispruner}, we use the first 1,000 samples from each benchmark for evaluation, and employ GPT-4.1 to assist in scoring open-ended responses. As shown in Table~\ref{tab:video_qa_results}, LearnPrunerV2 consistently outperforms FastV across all three benchmarks, especially under lower retention ratios. These results demonstrate the strong generalizability of LearnPrunerV2 from image-based to video-based VLMs.

\subsection{Data Efficiency}
\begin{figure}[!b]
    \centering
    \includegraphics[width=\linewidth]{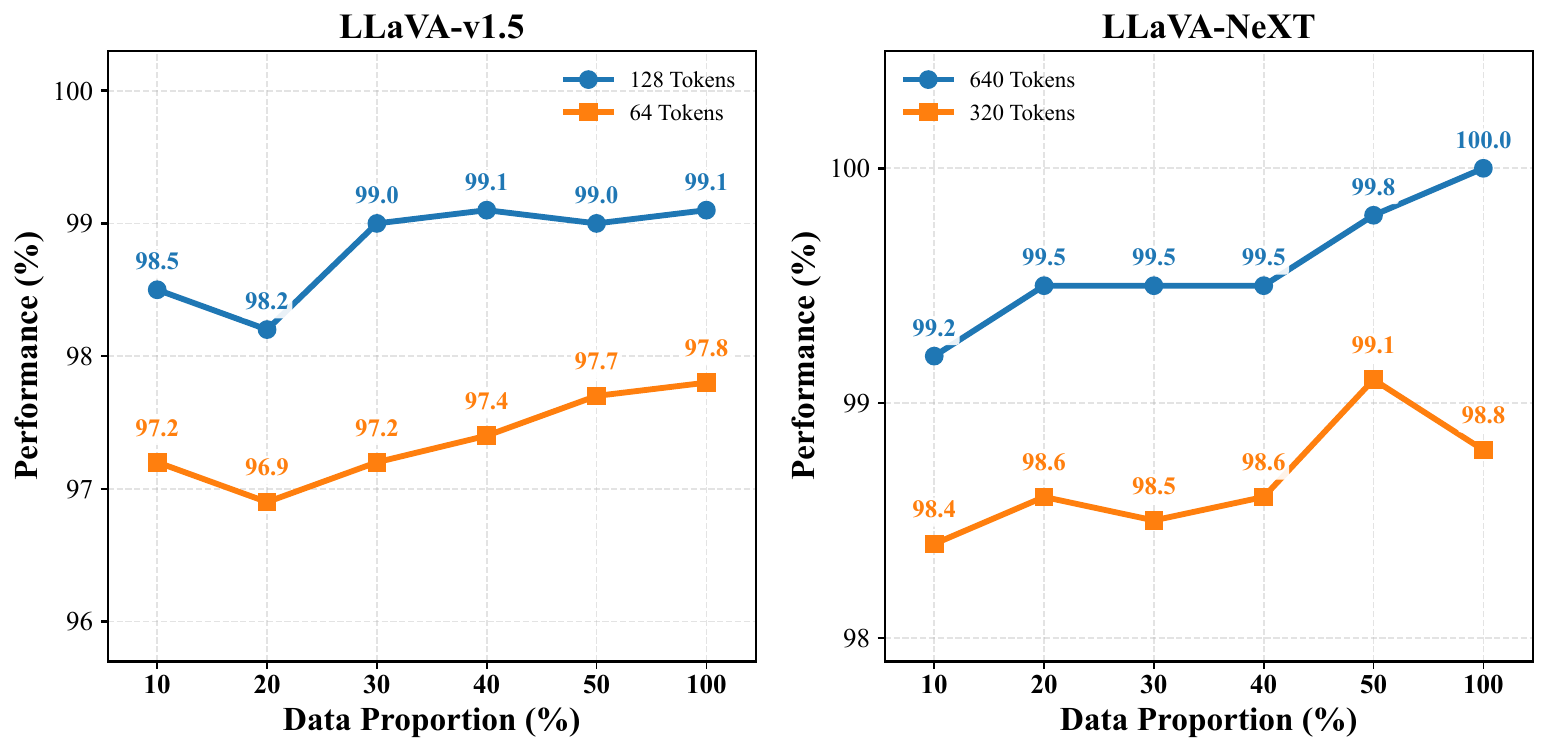}
    \caption{Performance comparisons of LearnPrunerV2 models trained with different proportion of the training dataset.}
    \Description{Enjoying the baseball game from the third-base seats. Ichiro Suzuki preparing to bat.}
    \label{fig:data_efficiency}
\end{figure}

To investigate the data efficiency of LearnPrunerV2, we train the model with varying proportions of the LLaVA-mix665k~\cite{liu2023visual} training dataset (from 10\% to 100\%)
and evaluate the resulting performance on both LLaVA-v1.5-7B and LLaVA-NeXT-7B
under two retention budgets, respectively. As shown in Figure~\ref{fig:data_efficiency}, LearnPrunerV2 exhibits remarkable data efficiency across both models. On LLaVA-v1.5, using only 10\% of the training data already achieves 98.5\% relative performance under the 128-token budget. As the training data scales to 50\%, performance largely converges, achieving results on par with full-dataset training. Moreover, we observe that lower token budgets generally require more training data to reach convergence. This can be attributed to the fact that more aggressive pruning demands a more precise pruning criterion, which in turn requires sufficient training signal to learn.

\subsection{Visualization}
\begin{figure*}[tbp]
    \centering
    \includegraphics[width=0.85\linewidth]{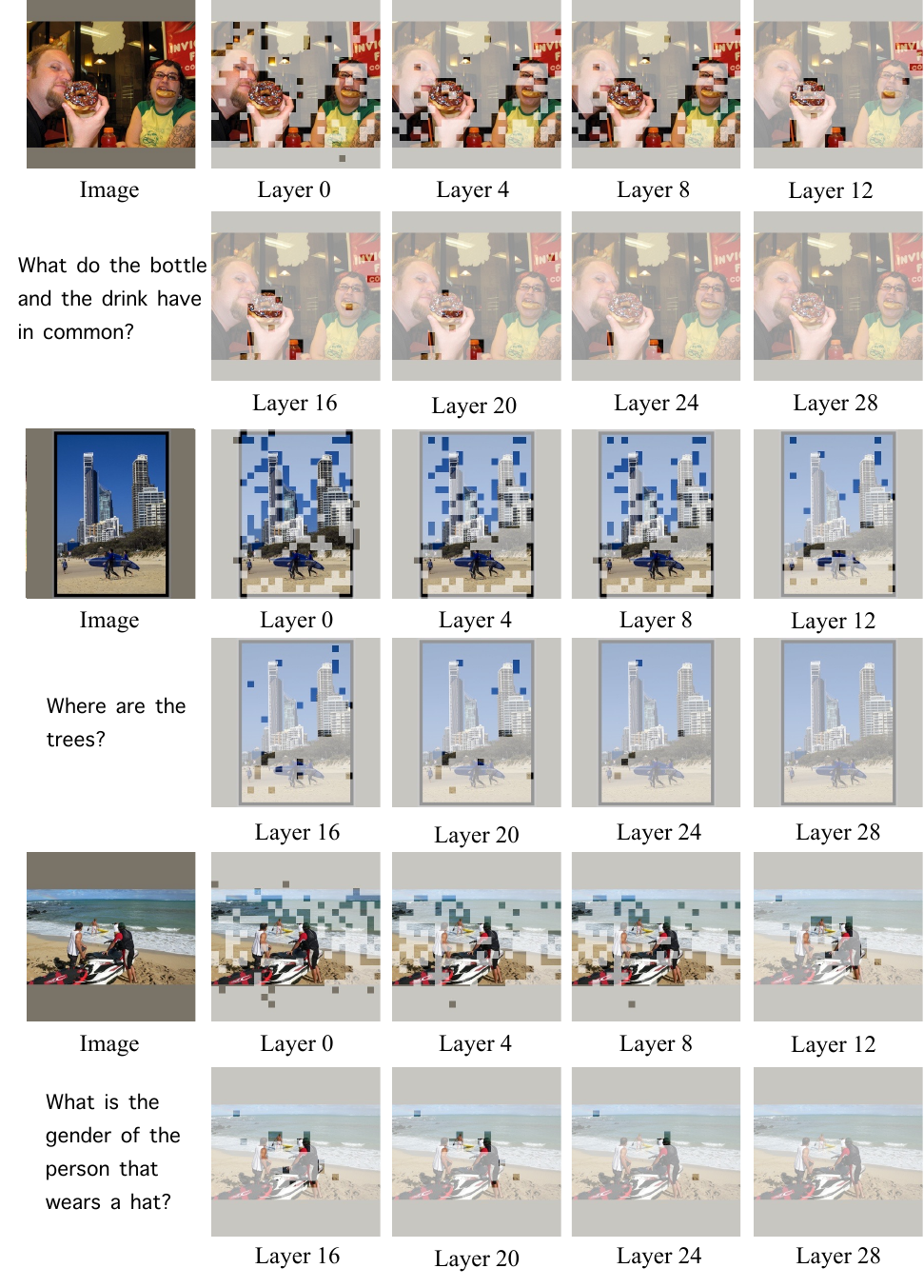}
    \caption{Visualization of the pruning results.}
    \Description{Enjoying the baseball game from the third-base seats. Ichiro Suzuki preparing to bat.}
    \label{fig:prune_vis_1}
\end{figure*}

\begin{figure*}[tbp]
    \centering
    \includegraphics[width=0.83\linewidth]{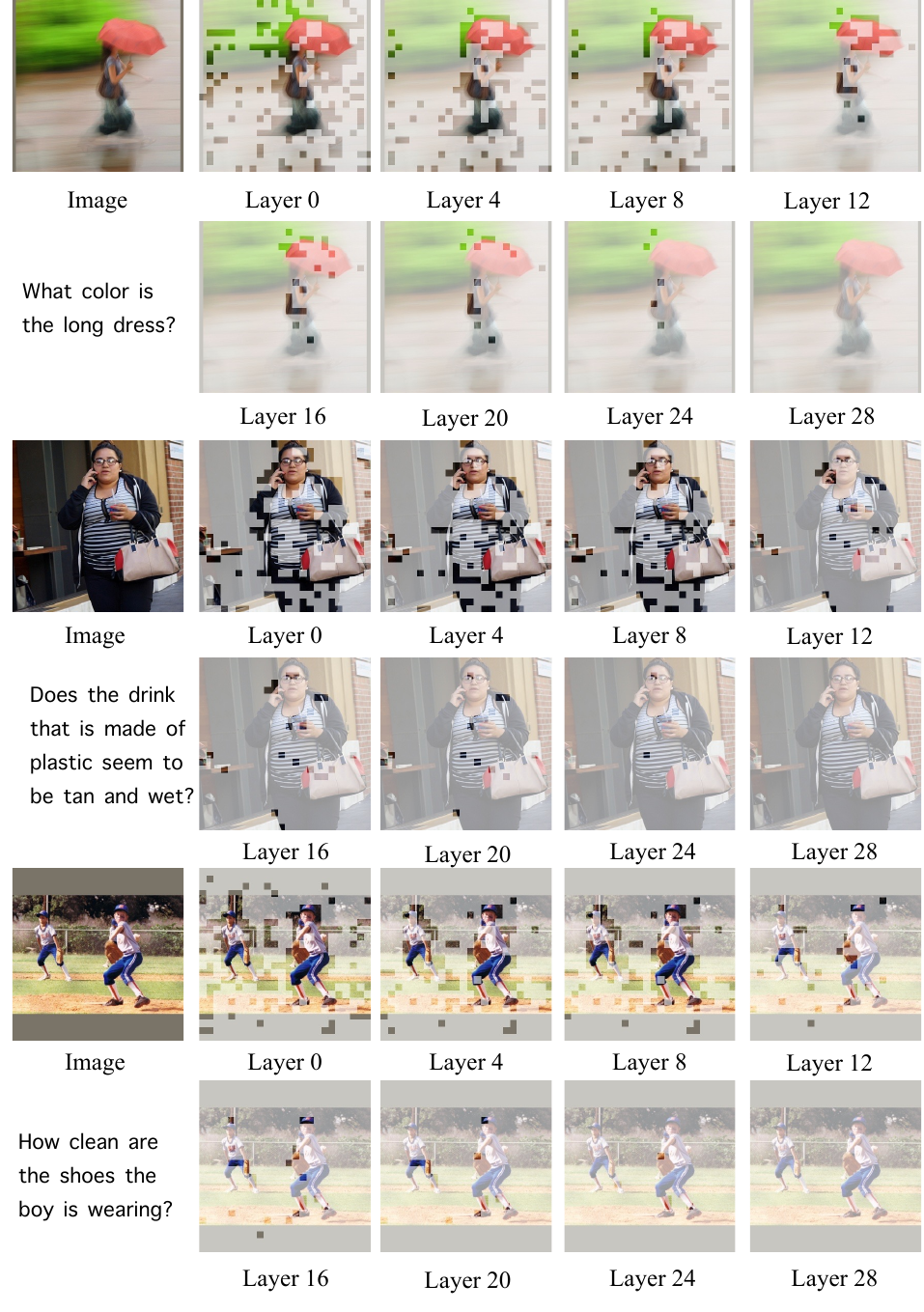}
    \caption{Visualization of the pruning results.}
    \Description{Enjoying the baseball game from the third-base seats. Ichiro Suzuki preparing to bat.}
    \label{fig:prune_vis_2}
\end{figure*}

To better illustrate the pruning behavior learned by LearnPrunerV2, we visualize the token pruning results across different layers on LLaVA-v1.5-7B. As shown in Figure~\ref{fig:prune_vis_1} and Figure~\ref{fig:prune_vis_2}, LearnPrunerV2 first removes a large number of uninformative tokens at shallow layers, such as those corresponding to padding regions or background areas. Moving to the middle layers, the retained tokens progressively converge toward query-relevant visual regions. By the deeper layers, nearly all visual tokens have been pruned. This progressive pruning pattern reveals the inherent characteristics of visual information flow within VLMs, which is consistent with findings in prior works~\cite{wu2026hidrop}.

\begin{figure*}[tbp]
  \centering
  \includegraphics[width=0.83\linewidth]{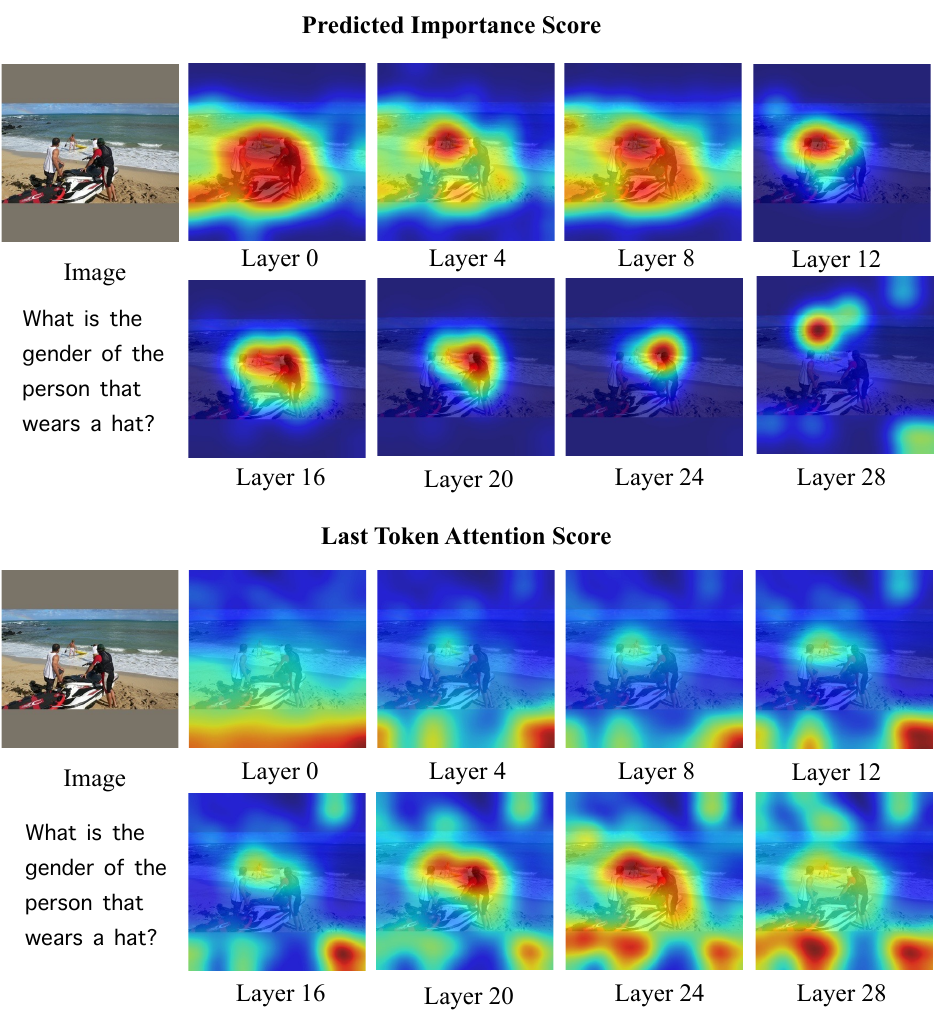}
  \caption{Visualization of the score heatmap.}
  \Description{Enjoying the baseball game from the third-base seats. Ichiro Suzuki preparing to bat.}
  \label{fig:score_vis_1}
\end{figure*}

\begin{figure*}[tbp]
  \centering
  \includegraphics[width=0.83\linewidth]{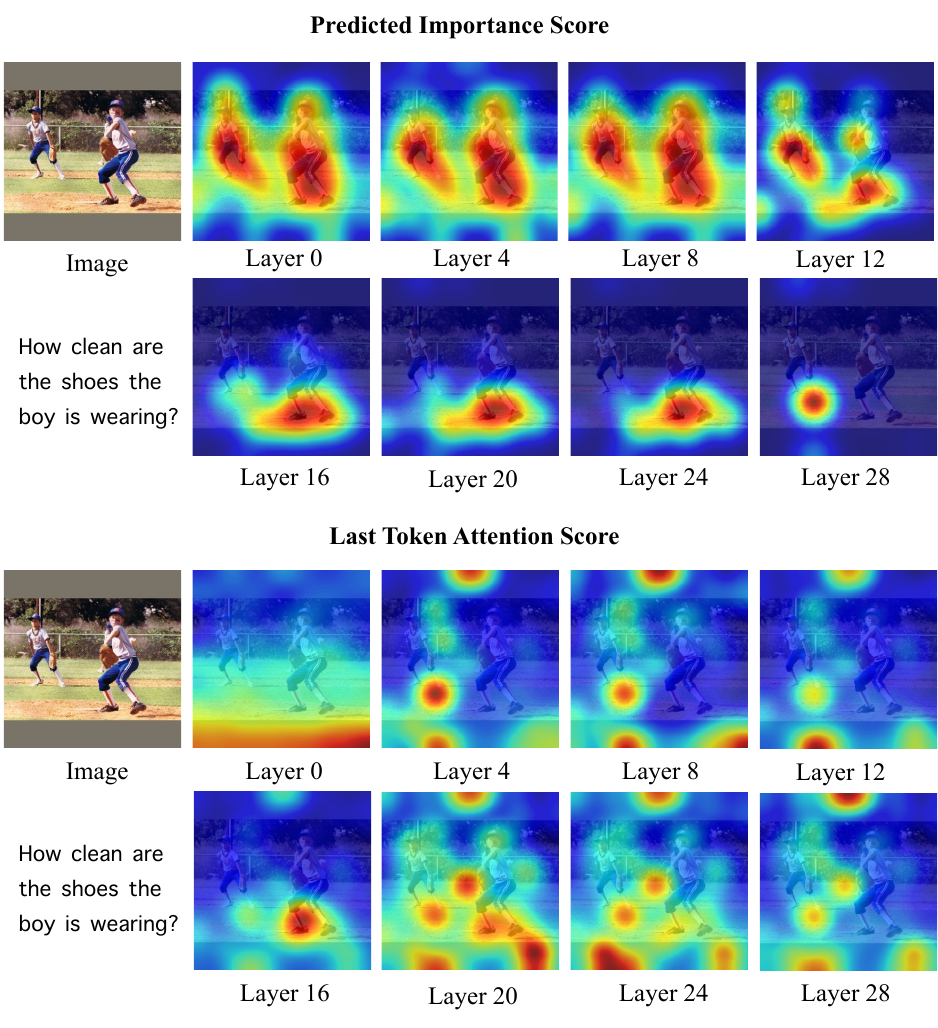}
  \caption{Visualization of the score heatmap.}
  \Description{Enjoying the baseball game from the third-base seats. Ichiro Suzuki preparing to bat.}
  \label{fig:score_vis_2}
\end{figure*}

Additionally, we compare the importance score heatmaps predicted by LearnPrunerV2 with the last-token attention scores adopted by previous methods. As shown in Figure~\ref{fig:score_vis_1} and Figure~\ref{fig:score_vis_2}, last-token attention suffers from notable positional bias and exhibits substantial fluctuations across layers, and typically requires deeper layers (e.g., layer 16) to localize query-relevant regions. In contrast, benefiting from our learnable shadow token and gating mechanism, LearnPrunerV2 is able to identify query-relevant regions at much earlier layers (e.g., layer 12), while also exhibiting stronger cross-layer consistency. 

\begin{table}[!b]
  \centering
  \small
  \begin{tabular}{l|cc|cc|cc}
  \toprule
  \multirow{2}{*}{\textbf{Method}} & \multicolumn{2}{c|}{\textbf{TGIF-QA}} & \multicolumn{2}{c|}{\textbf{MSVD-QA}} & \multicolumn{2}{c}{\textbf{MSRVTT-QA}} \\
  & \textbf{Acc.} & \textbf{Score} & \textbf{Acc.} & \textbf{Score} & \textbf{Acc.} & \textbf{Score} \\
  \midrule
  \rowcolor{gray!30}
  \multicolumn{7}{c}{\textit{Upper Bound, 2,048 Tokens (100\%)}} \\
  Vanilla & 18.1 & 1.27 & 64.1 & 3.41 & 56.1 & 2.97 \\
  \midrule
  \rowcolor{gray!30}
  \multicolumn{7}{c}{\textit{Retain Averaged 456 Tokens (\textcolor{blue}{$\downarrow$ 77.8\%})}} \\
  FastV & 16.4 & 1.17 & 60.3 & 3.24 & 53.0 & 2.84 \\
  LearnPrunerV2 & \textbf{17.8} & \textbf{1.21} & \textbf{64.9} & \textbf{3.39} & \textbf{55.0} & \textbf{2.89} \\
  \midrule
  \rowcolor{gray!30}
  \multicolumn{7}{c}{\textit{Retain Averaged 228 Tokens (\textcolor{blue}{$\downarrow$ 88.9\%})}} \\
  FastV & 12.4 & 1.00 & 55.1 & 2.93 & 50.4 & 2.70 \\
  LearnPrunerV2 & \textbf{16.4} & \textbf{1.19} & \textbf{62.1} & \textbf{3.30} & \textbf{53.8} & \textbf{2.84} \\
  \bottomrule
  \end{tabular}
  \caption{Video understanding performance comparisons of different pruning methods on Video-LLaVA-7B. Performance is evaluated on the first 1,000 samples of each benchmark. We employ GPT-4.1 to assist in scoring open-ended responses.}
  \vspace{-0.25cm} 
  \label{tab:video_qa_results}
\end{table}

\end{document}